\documentclass[sigconf]{acmart}

\copyrightyear{2022} 
\acmYear{2022} 
\setcopyright{acmlicensed}
\acmConference[AIMLSystems 2022]{The Second International Conference on AI-ML Systems}{October 12--15, 2022}{Bangalore, India}
\acmBooktitle{The Second International Conference on AI-ML Systems (AIMLSystems 2022), October 12--15, 2022, Bangalore, India}
\acmPrice{15.00}
\acmDOI{10.1145/3564121.3564126}
\acmISBN{978-1-4503-9847-3/22/10}

\usepackage{booktabs}       
\usepackage{microtype}      
\usepackage{xcolor}         
\usepackage{graphicx}
\usepackage{algpseudocode}
\usepackage{algorithm}
\usepackage{array}
\usepackage{cleveref}

\graphicspath{ {./images/} }

\begin{document}

\title{Unposed: Unsupervised Pose Estimation based Product Image Recommendations}

\author{Saurabh Sharma}
\affiliation{%
  \institution{Amazon.com}
  \city{Bengaluru}
  \country{India}
}
\email{sharsar@amazon.com}

\author{Faizan Ahemad}
\affiliation{%
  \institution{Amazon.com}
  \city{Bengaluru}
  \country{India}
}
\email{ahemf@amazon.com}

\renewcommand{\shortauthors}{Sharma et al.}

\begin{abstract}
  Product images are the most impressing medium of customer interaction on the product detail pages of e-commerce websites. Millions of products are onboarded on to webstore catalogues daily and maintaining a high quality bar for a product's set of images is a problem at scale. Grouping products by categories, clothing is a very high volume and high velocity category and thus deserves its own attention. Given the scale it is challenging to monitor the completeness of image set, which adequately details the product for the consumers, which in turn often leads to a poor customer experience and thus customer drop off.
   
  To supervise the quality and completeness of the images in the product pages for these product types and suggest improvements, we propose a Human Pose Detection based unsupervised method to scan the image set of a product for the missing ones. The unsupervised approach suggests a fair approach to sellers based on product and category irrespective of any biases. We first create a reference image set of popular products with wholesome imageset. Then we create clusters of images to label most desirable poses to form the classes for the reference set from these ideal products set. Further, for all test products we scan the images for all desired pose classes w.r.t. reference set poses, determine the missing ones and sort them in the order of potential impact. These missing poses can further be used by the sellers to add enriched product listing image. We gathered data from popular online webstore and surveyed \textasciitilde200 products manually, a large fraction of which had at least 1 repeated image or missing variant, and sampled 3K products(\textasciitilde20K images) of which a significant proportion had scope for adding many image variants as compared to high rated products which had more than double image variants, indicating that our model can potentially be used on a large scale.
\end{abstract}



\begin{teaserfigure}
  \includegraphics[width=\textwidth]{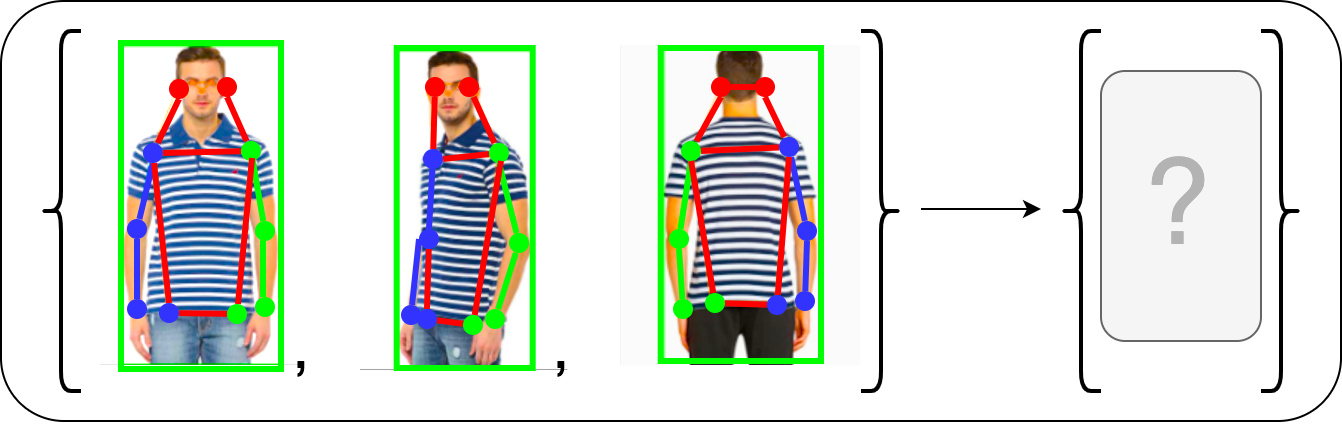}
  \caption{Pose detection on Product images to propose images.}
  \label{fig:teaser}
\end{teaserfigure}

\maketitle

\section{Introduction}
On the e-commerce stores, the product information is presented to the customer primarily in the 3 modes: \emph{text, video and image}. An accurate and wholesome presentation of the product details is critical to influence purchase decision of the customer. With more and more sellers offering an even wider selection of products, it becomes unscalable to maintain the catalogue quality by checking for missing information.

Depending on the domain of the business the 3 modes of presenting product details may have different impact. For example, a customer purchasing an electrical component may be more interested in detailed specifications in text format, whereas the one, a winter jacket, may be interested in visual image of the product. Even for image friendly categories, given the variety and nuances of products, a one-size-fits-all solution for absent images is ineffective. So we focus on fast moving Apparel product type and outline a solution to suggest missing images in the following steps:
\begin{enumerate}
   \item Determine ideal imageset in an unsupervised manner.
   \item Scan all products of the category for missing images against this set.
   \item Propose the image candidates ordered by potential impact.
\end{enumerate}

\subsection{Catalogues in clothing}
\label{sec:intro1}
Clothing product pages tend to show images of human models / mannequins exhibiting the clothing / accessories from different poses / directions. However, many products on the webstores may suffer from poor cataloguing practices like duplicated images, bad image lighting, poor background, insufficient number of angles of clothing, etc. leading to poor customer experience. A few A/B experiments 
done at Amazon have confirmed this and have shown a non-trivial positive impact of having larger number and varied type of images.

The problem of inadequately imaged product, thus has a user experience improvement incentive. Any automation that can meaningfully nudge the selling partners with right suggestions for improving their catalogue images will help them better their listing. This in turn can improve the end user experience, there by driving the virtuous cycle.

\begin{figure*}[h]
   \centering
   \includegraphics[width=0.8\linewidth]{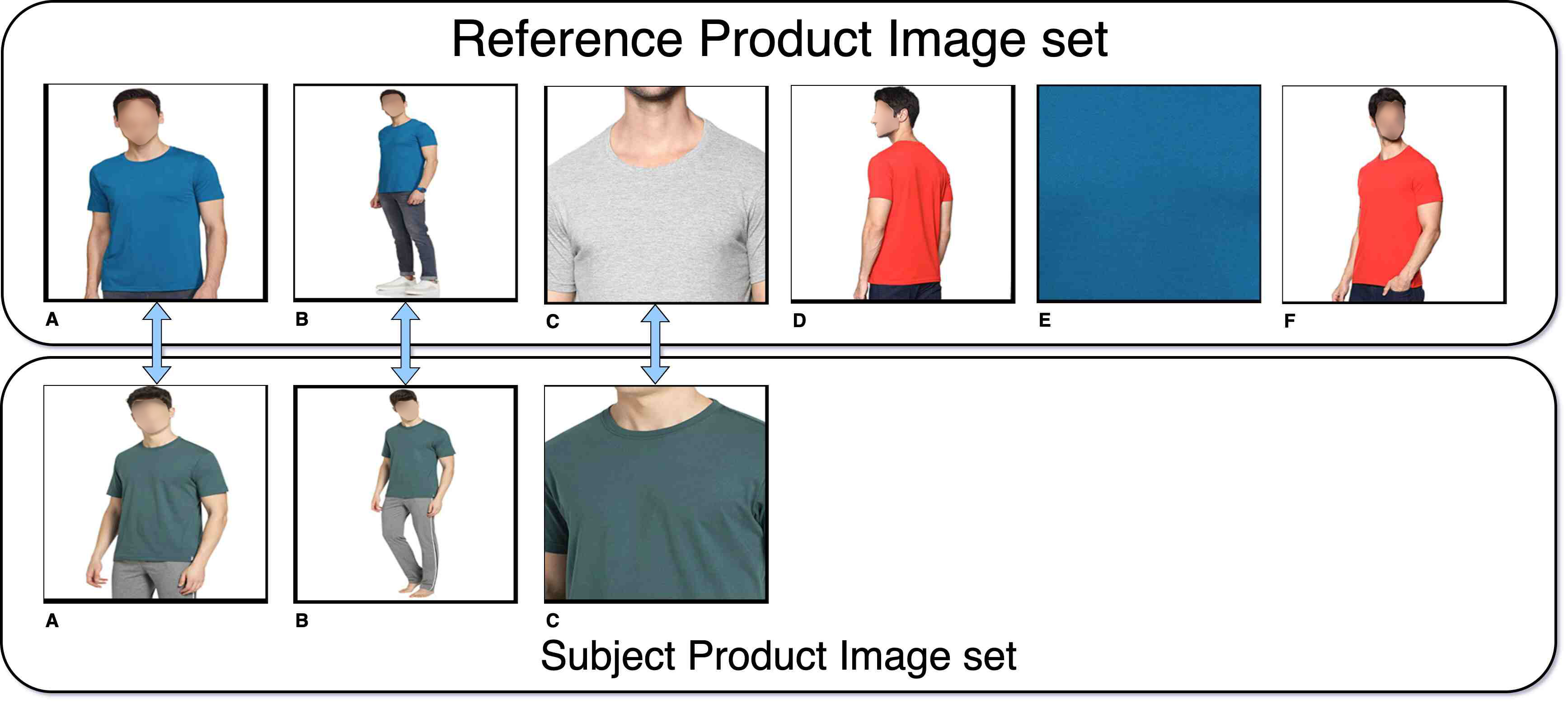}
   \caption{Reference vs Subject product image set comparison $\mid$  $\S$\ref{sec:intro1}}
   \label{fig:compare}
\end{figure*}

The missing images can be indicated to the sellers in text hints, or they can be provided reference image set of popular products in the Category/ Subcategory corresponding to their underrepresented product in the clothing category. For example, Figure \ref{fig:compare} shows the \emph{reference} imageset with 6 images, covering \emph{front, farview, neckview, back, fabric pattern, side view} etc. whereas, the subject product image are missing a few from the reference set. In this work, we discuss ways to detect, and indicate the missing images to improve the product listing by detecting missing poses in the image set.

\section{Related Work}
This section discusses prior work done for catalogue quality improvement as well as gives a brief refresher on the problem of human pose detection.

\subsection{Catalogue Image quality}
The product listing image selection is largely done manually by professionals specializing in product listing adhering to Standard Operating Procdeures based on listing guidelines. Some prior published work \cite{smartimageselection} suggests solutions to build systems to solve image listing problems like de-duplication, image cropping etc. 

In our scheme we suggest what images can be added to the Product imageset to enhance the user experience, and no prior work in our knowledge discusses pose detection (discussed next $\S$\ref{sec:PosedDetection}) as a method to generate candidates for image addition. 

\subsection{Pose Detection}
\label{sec:PosedDetection}
Human pose detection has been a widely studied computer vision application. The core problem of human pose detection is to identify the body landmarks or regions/points of interests (PoI) like \emph{nose, shoulder, knee} etc. and predict coordinates of these landmarks \emph{w.r.t.} the given image. A few prominent models use HourGlass\cite{Hourglass} and U-NET\cite{unet} type of Encoder-Decoder NN architecture to identify Points of Interest.

Then there are multistage approaches where in the Bottoms-Up approach the model first detects body joints location on the image and then links them to guess complete human instance. In Top-down approach models \cite{MSEHPN}, in the first stage the model identifies the human body in the image, and in subsequent stages identifies landmarks. The current state-of-the-art model is based on Fully Convolutional Networks\cite{SGFCN}, with PCKh (87\%) on the MPII dataset \cite{MPII}. We use pretrained \emph{MMPose} \cite{mmpose2020} toolbox library which implements HRNet full body keypoints detection model \cite{HRNet_2019_CVPR} and \emph{Blazepose}\cite{BlazePose}, which is a lightweight implementation of HourGlass style network architecture to generate position embeddings in our pipeline.

\section{Unpose}
The Unpose system presents an unsupervised method for classifying, and ranking most popular poses, and further extends the framework to detect the missing images from target products. The system attempts to:
\begin{enumerate}
   \item Define the ideal set of images in a reference set.
   \item Detect the missing images from the subject imageset \emph{w.r.t.} to the reference set.
   \item Rank the identified images in decreasing order of importance.
\end{enumerate}
The test product \emph{subject} imageset mentioned here has less than p-threshold($p$=5) number of images in the listing and is our target for improvement. We discuss this scheme's components in following sections starting by building the reference imageset.

\subsection{Building reference imageset}
We built the reference imageset by selecting top-K product listings by popularity belonging to \emph{Clothing} sorted by number of customer average review ratings and number of reviews. This sorting criteria is easily configurable based on available signals.
\subsection{Reference set processing flow}
\label{sec:trainflow}
\begin{figure*}
   \begin{center}
    \includegraphics[width=0.8\linewidth]{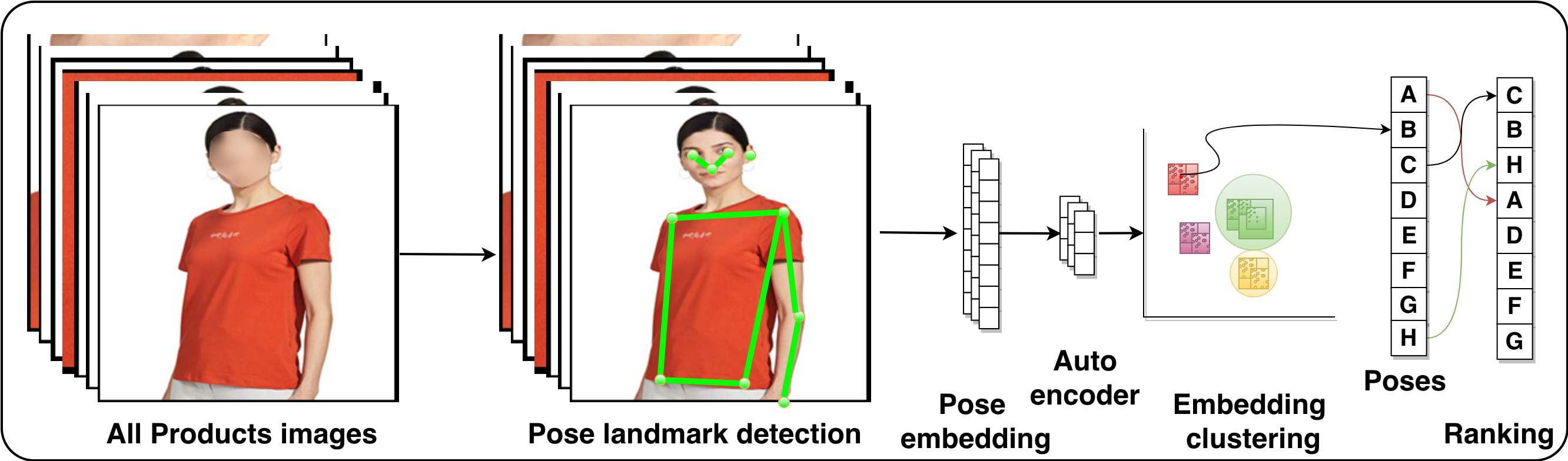}
   \end{center}
      \caption{Training pipeline for missing image detection $\mid$ $\S$\ref{sec:trainflow}}
   \label{fig:short}
   \end{figure*}

The reference set images selected like described in the previous section are then passed through a data pipeline (Figure \ref{fig:short}) which
\begin{enumerate}
   \item Reads the image.
   \item Estimates the coordinates of various human body landmarks like \emph{nose, shoulders, knees, wrists, elbows} etc. w.r.t. the image. 
   \item Normalizes each of these coordinates w.r.t. to image size.
   \item Creates a vector representing one image from the imageset using the normalized coordinates.
   \item Creates clusters of the vector representation of all the images using K-Means algorithm to detect K centroids, where each centroid corresponds to cluster of similar poses detected in all the images in the train imageset.
   \item Ranks all the centroids for each cohort grouped by attributes like category, subcategory, product type based on occurrence in most popular / best rated products to determine importance of each centroid within the cohort.
\end{enumerate}

\subsection{Detecting the missing images in Test Set}
\label{sec:testflow}
For a given product, all the images are collected as a subject product imageset. To detect the missing images from this subject imageset w.r.t. reference
imageset:
\begin{enumerate}
   \item The model first generates the landmarks / pose embeddings coordinates for all images using the same estimation method as reference dataset.
   \item Then after normalizing these coordinates w.r.t. to image size, the vector embeddings are created.
   \item For each image, the model then finds the nearest centroid corresponding to approximate closest pose w.r.t. reference poses detected in the previous section.
   \item Further each centroid for which there were no images in the subject imageset, is ranked as per popularity criteria defined while training the reference set. Then these missing poses are flagged in the decreasing order of popularity or usefulness as illustrated in Figure \ref{fig:sample}.
\end{enumerate}

\subsection{Proposed image candidates}
After finding the indices / labels corresponding to missing centroid in the inference flow mentioned in previous section, the user can refer the reference imageset to find out the reference images corresponding to these centroids for a given Product. 

\subsection{Formulation}
Given a set $\mathbf{Q_D} = \{q_1^D, q_2^D..., q_n^D\}$ of $n$ images for a subject product $\mathbf{D}$ in imageset and a universal set $\mathbf{U} = \{u_1, u_2,..u_m\}$ of all the $m$ images of Top-K products in the training product imagesets, the problem is to determine a set $\mathbf{T}$ of tags, $1 \leq i \leq o$, corresponding to $\mathbf{o}$ optimal clusters $C_1, C_2, ..., C_o$ for images from $\mathbf{U}$ and further find a set $t_j \subseteq  \mathbf{T}$, $1 \leq j \leq o$ of tags, corresponding to images in subject imageset $q_{t_j}^D\mid q_{t_j}^D:q_{t_j}^D \not\in \mathbf{Q_D}, 1 \leq j \leq o$.

To calculate these we first determine the ideal $\mathbf{o}$-clusters from the images in $\mathbf{U}$ representing all images of all Product in reference training sets. 
Let \{$x_1, x_2,..., x_G\} \in \mathbf{X_v}$ be the vector representation of any image in $\mathbf{U}$, then we determine $\mu_o$ vectors representations of these $\mathbf{o}$-clusters such that we minimize:

$$ \sum_{k=0}^{m}\sum_{i=0}^{G}{\min_{j \in \{0,o\}}(\parallel x_i - \mu_j\parallel^2_m)}$$
Here the vector $\mathbf{X_v}$ is created by using normalized position coordinates of Point of Interests (PoI) from Pose Detection.

Later we determine all the tags $l_j$ so as to maximize the probability $p(q_i\mid \mathbf{X_v})$ of image $I$ to belong to the cluster given by tag
   $l_j \in \mathbf{T}$ such that $$j = argmin{(\parallel x_i^I - \mu_j\parallel^2_I)}\ \forall\ x_i^I \in \mathbf{X_v^I}$$

Further we determine all tags $t_j$'s such that, all tags that are in T but not are detected in the previous step.
$$ t_j = \{t_i : t_i \in \mathbf{T},\ t_i \not \in {l_i} \forall\ 1 \leq i \leq o\}$$
These $t_j$'s will be the missing poses / centroid in the subject imageset w.r.t. $\mathbf{o}$ centroids defined from reference imageset.

\section{Pipeline}
The training pipeline that consists of Pose landmark detection, Embedding generation, the K-means clustering for ideal centroid detection, followed by ranking mechanism is described in the text below. We focus on the dataset creation used in the pipeline, then proceed to individual stages of processing. 

\subsection{Dataset}
We aggregated public data from popular e-commerce website's clothing products metadata and listing images, and selected top 3000 most popular products with available image count per product > 9 based on customer ratings, etc. This collection provided us with an imageset of \textasciitilde20K for training. For evaluation purpose, we selected unseen products from the similar selection criteria collection of top products, but such that images per product imageset were < 5. 

\subsection{Pose Landmarks}
\begin{figure}[H]
      \centering
      \includegraphics[width=0.5\linewidth]{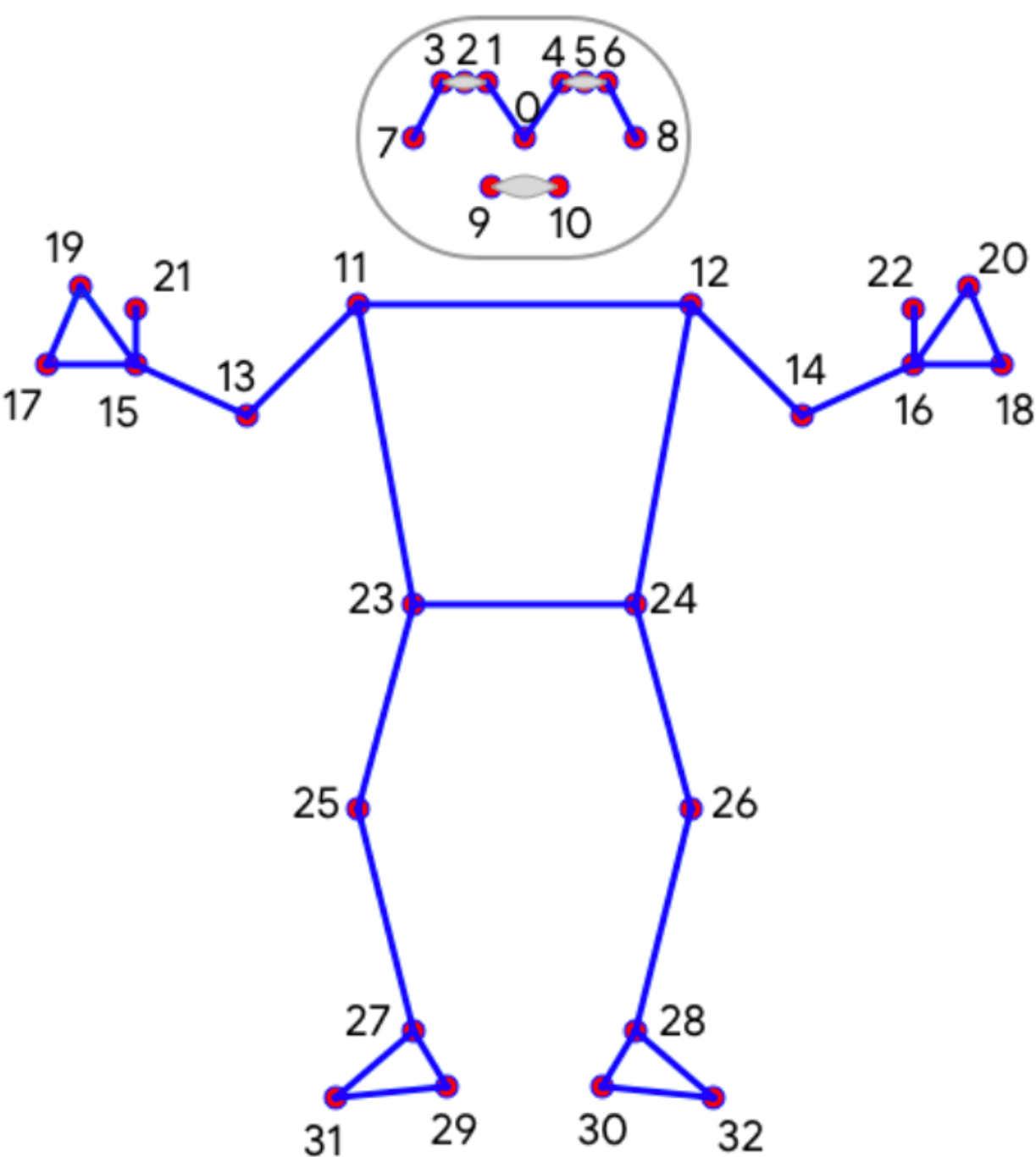}
      \caption{BlazePose landmarks\cite{BlazePose}}
      \label{fig:blazepose}
   \end{figure}

For experimentation, the Pose Landmark coordinates are estimated using \emph{Blazepose} pretrained implementation of HourGlass style Region detection NN library as well as \emph{MMPose} toolkit pretrained implementation of HRNet. The implementation Blazepose \cite{BlazePose} (MMPose \cite{mmpose2020}) generates 3D coordinates for 33 (HRNet 2D: 17) landmarks points as depicted in Figure~\ref{fig:blazepose}. These coordinates are normalized w.r.t. images dimensions. The pose landmarks are cartesian pixel coordinates w.r.t. to image top left.

\begin{figure*}
   \begin{center}
    \includegraphics[width=\linewidth]{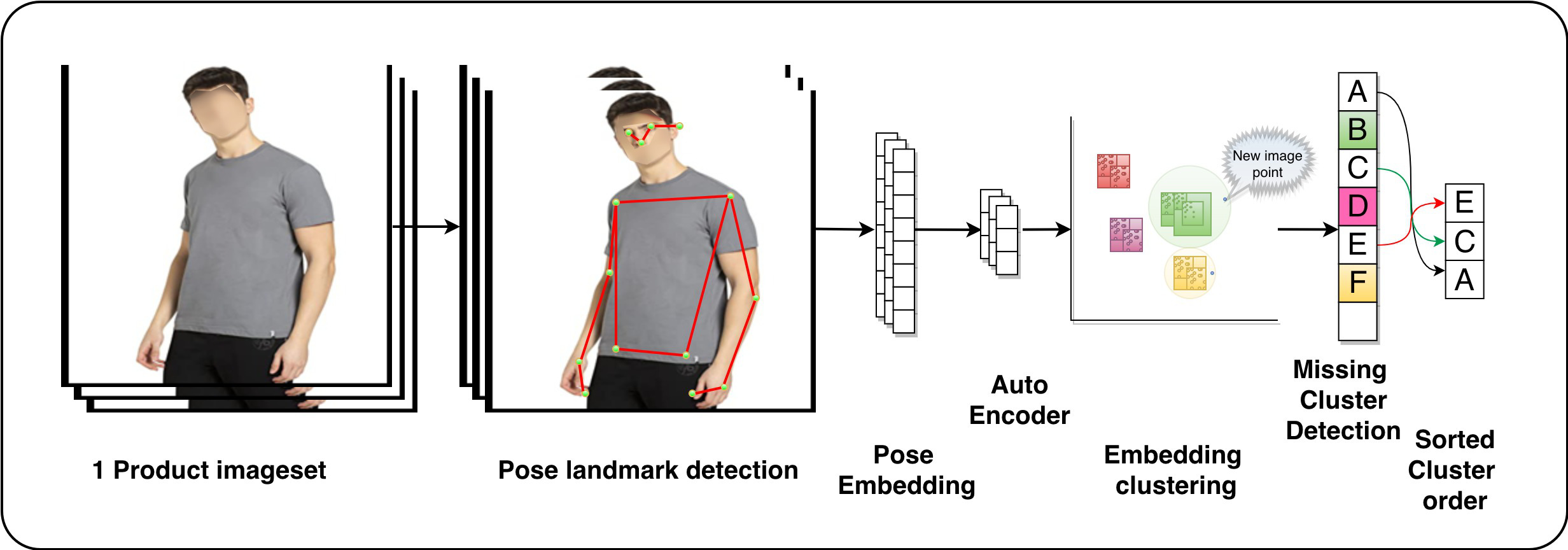}
   \end{center}
   \caption{Inference pipeline for missing image detection $\mid$ $\S$\ref{sec:inference}}
   \label{fig:inference}
\end{figure*}

\subsection{Embeddings}
The embedding vectors are compiled using landmark coordinates w.r.t. to image size. To add more pose related information, the embeddings are extended by taking various ratios of \emph{x}, \emph{y} coordinates of landmarks to detect the front, side, backview, e.g. $\tfrac{\mathbf{x}_{left\ shoulder}} {\mathbf{x}_{right\ shoulder}}$. Further the \emph{z} components (available only for 3D-model) help identify if the image is close-up, distant pose etc. We generate a 77 (MMpose2D: 61) dimensional vector to store the features thus engineered. These embeddings are calculated for all training as well as test set images.

\subsubsection{Autoencoder Embeddings}
Further to develop efficient representations, the training imageset pose embeddings are fit and transformed through an \emph{Autoencoder} to reduce the dimensionality of the embeddings, and thus learn a compressed representation of raw data to observe an improvement in clustering performance. The autoencoder optimizes for embeddings by minimizing for batchwise contrastive loss. 

The Autoencoder we used is a encoder-decoder network which optimizes contrastive loss of 2048 batch size. The encoder is 4 layer fully connected feed forward network (FCFFN) with input dimensionality of embeddings from previous layer and bottleneck size of 8 nodes. The decoder is a 2 layer FCFFN.  It is trained using AdamW optimizer with decay of 0.001 and starting learning rate of 0.1. 

\subsection{Clustering}
After generating the embeddings, the pipeline further uses clustering using K-Means implementation of faiss \cite{faiss} library to identify K-centroids which represents mean value of K vectors over the training image set embeddings. The centroid and clusters thus generated signify the embeddings corresponding to the most similar and recurring poses in reference image dataset. The clustering algorithm is fit over flattened representations of all images in the training imageset.  

\subsection{Ranking the poses}
After calculating the centroids for the clusters, we determine the suitability order of each of the centroids, which in turn correspond to the poses. To quantify this suitability, for each image in the reference dataset, we estimate the closest pose (represented by the index value of cluster centroid, a value $[0, Num_{poses})$), and then rank these centroids against the target variable which is weighted mean of image's Product's customer average review rating, number of ratings etc. To account for the effect of category, subcategory, and other categorical attributes, we train a GBDT based regressor with these features along with centroids to predict the target variable. This gives us a relative order of a centroid for a given category, subcategory, etc. The Algorithm \ref{alg:training} summarizes the scheme.

\subsection{Inference}
\label{sec:inference}
The inference pipeline (Figure \ref{fig:inference}) follows the same process of PoI detection, embedding generation. However, inference is calculated at per Product listing - imageset. The pipeline calculates the index of nearest centroid corresponding to each image in the imageset. The inference flow is outlined in the   
 Algorithm \ref{alg:inference}.

\begin{algorithm}
   \begin{algorithmic}[]
      \State AllCentroids $\gets \{\}$
      \State AllImageVector $\gets [\ ]$
      \State RankVector $\gets$ None
      \For{$i$ = 0 to len(AllTrainingImages)}
         
      \State landmarks $\gets$ \Call{GetLandmark}{AllTrainingImages[$i$]}
         
      \State embedding $\gets$ \Call{GetEmbedding}{landmarks}
      \State embedding, AutoencVector $\gets$ \Call{Autoenc}{embedding}
   
      \State {AllImageVector[$i$] $\gets$ embedding}
      \EndFor
      \State AllImageVector, AutoencVector $\gets$ \Call{Autoenc}{AllImageVector}
      \State AllCentroids $\gets$ \Call{CalcAllCentroids}{AllImageVector}
      \State RankVector $\gets$ \Call{LearnRank}{AllCentroids, AllTrainingImagesData}
   
      \Return AllCentroids, RankVector, AutoencVector
   
   \end{algorithmic}
   \caption{Training flow $\mid$ $\S$\ref{sec:inference}}
   \label{alg:training}
   \end{algorithm}

\begin{algorithm}
\begin{algorithmic}[]
   \State haveCentroids $\gets$ \{\}

   \For{$i$ = 0 to len(ImageSet)}
      
   \State landmarks $\gets$ \Call{GetLandmark}{ImageSet[$i$]}
      
   \State embedding $\gets$ \Call{GetEmbedding}{landmarks}
   \State embedding $\gets$ \Call{Autoenc}{AutoencVector, embedding}
   \State centroid $\gets$ \Call{GetNearestCentroid}{embedding}
   \State {haveCentroids $\gets$ haveCentroids\, ||\, centroid}
   \EndFor

   \State MissingCentroids $\gets$ \{\}
   
   \For{$i$ = 0 to len(AllCentroids) }
   \If{$i$ is not in haveCentroids}
       \State {MissingCentroids $\gets$ MissingCentroids\, ||\, $i$}
   \EndIf
   \EndFor
   \State MissingCentroids $\gets$ \Call{SortCentroidsByRank}{RankVector, MissingCentroids, ProductMetatData} 

   \Return MissingCentroids

\end{algorithmic}
\caption{Inference flow $\mid$ $\S$\ref{sec:inference}}
\label{alg:inference}
\end{algorithm}

The MissingCentroids thus obtained gives the indices of the Centroids corresponding to the pose w.r.t. AllCentroids representing the reference set centroids and are determined during the training phase. In Figure \ref{fig:sample} we observe that $\{D,E,F\}$ are the detected as the missing Centroids in decreasing order of importance, and user can refer the corresponding images from the reference set.

\setcounter{figure}{5}
\begin{figure*}
      \centering
      \includegraphics[width=\linewidth]{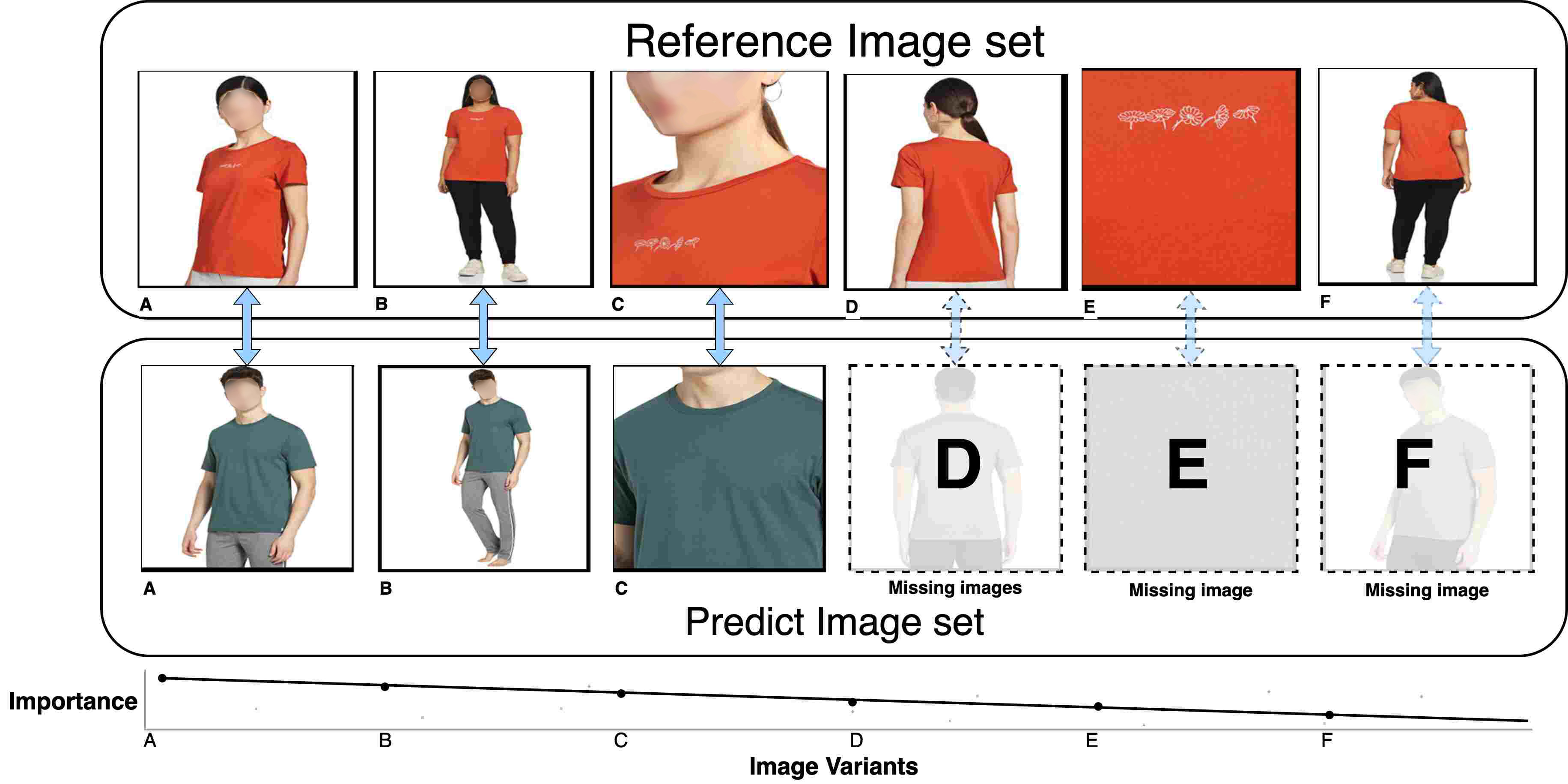}
      \caption{Sample imagesets $\mid$ $\S$\ref{sec:testflow},  $\S$\ref{sec:inference}\\
 The figure shows how the ideal \emph{Reference} imageset has centroids \textbf{A-F} and in the \emph{Predict} imageset only centroids \textbf{A-C} are present, with MissingCentroids indicating $\{D, E, F\}$ to be the missing images in decreasing order of importance. Please refer Appendix appendix \ref{sec:appendix} for more examples.}
   \label{fig:sample}
\end{figure*}

\section{Performance and Evaluation}
\label{sec:perf}
The performance of the model should be evaluated on at least 2 parameters:
\begin{enumerate}
   \item \label{itm:first} \textbf{Quality of clusters generated} - Are the clusters generated for true images cover all variations of images as per cataloguing guidelines.
   \item \label{itm:second} \textbf{Accuracy of clustering of the validation set} - As compared to human labelled images, is clustering enumerating the missing images correctly or not. This in turn depends on the definition of embedding quality.
\end{enumerate}

Item \ref{itm:first} is not quantitatively assessed in the text, however few examples of cluster labels are presented for subjective evaluation in  Table \ref{table:centroiddescription}. This can be addressed quantitatively in future work based on aesthetic evaluations.
Item \ref{itm:second} is assessed based on manual labelling the validation set images. This is achieved by comparing imageset cluster labels predicted by the model vs. manually labelling an image w.r.t. cluster label of most similar image with similar pose from training set images.
\begin{table}[h]
   \centering
   \caption{Variant label model accuracy on different subcategory under Clothing category $\mid$ $\S$\ref{sec:perf}}
   \label{table:accuracy}
   \begin{tabular}{rlp{0.2\linewidth}p{0.2\linewidth}}
      \toprule
       & \multicolumn{2}{c}{BlazePose}&MMPose\\ 
   \textbf{Subcategory} & \textbf{Accuracy} & \textbf{Accuracy [Autoenc]}& \textbf{Accuracy [Autoenc]}\\
   \midrule
   Polo shirts          & 0.6967 & 0.8264          & \textbf{0.8536}    \\ 
   T-shirts             & 0.9186 & \textbf{0.9652} & 0.9565             \\ 
   Men casual shirts    & 0.8442 & 0.8823          & \textbf{0.91089}   \\ 
   \bottomrule      
   \end{tabular}

\end{table}

where the Accuracy is defined as:

$$Accuracy = \frac{Total\ num.\ of\ Labels\ detected\ missing\ by\ model}{True\ num.\ of\ Labels\ missing\ as\ per\ human\ label}$$

The Table \ref{table:accuracy} shows accuracy numbers for PoI coordinates based embedding \emph{Accuracy} and PoI coordinates along with Autoencoder reduced embeddings \emph{[Accuracy[Autoenc]]} for BlazePose implementation of the flow. The column MMPose highlights the Accuracy numbers when embeddings were generated using MMPose2D pose estimation PoI coordinate values. The high accuracy numbers may be misleading because validation set imagesets are having very few images / repeated images, hence large num of centroids are recommended as missing, however this imageset-wise accuracy does count for correct label assignment to each individual image and is certainly indicative of effectiveness of the approach. Each subcategory set in the Table \ref{table:accuracy} had 20 product image sets with 4 or fewer images.

\section{Model implementation notes}
\label{sec:notes}
We observed a few limitations in the current work with the implementation which may or may not be influence business use case, they are as follows:
\begin{itemize}
   \item Blazepose model is not able to output non-trivial coordinates when either of limbs or head is not visible, but MMPose is able to determine coordinates if some boundary structure is available. E.g. in a t-shirt image if neck and above region cropped MMpose is able to identify it as pose. Refer  Figure \ref{fig:tshirte2} in Appendix.
   \item Both Blazepose and MMPose models are not able to distinguish between male and female posers. This is generally not a con as it is seldom required to have both male/females poses in the image set.
   \item Behavior of model is unpredictable when there are more than one posers/apparels in the image.
   \item These models classify all non-human image under 1 centroid. E.g. size-chart, fabric view etc. will be grouped as same pose. 
\end{itemize}

\section{Conclusion}
The text discusses a novel and promising approach to recommend images to populate image in listings for a high volume and popular category. This model can be one of the many approaches required to address different aspects like catalogue images with \emph{pose, lighting, color, crop} etc. issue in the image listings. To improve the coverage and accuracy of the model, the processing and data set generation can be improved by adequate data exploratory analysis. The mechanism suggested is free of human bias in the dataset if any, and recommendations thus generated are based on statistics of end user preferences, and product offering by the seller and is easily customizeable for any e-commerce website, given available levels of telemetry data.

\balance
\bibliographystyle{ACM-Reference-Format}
\bibliography{PoseDetectionAIMLSys.bib}

\appendix

\section{Sample outputs}
\label{sec:appendix}
This section shows some experiment results samples.

Since Unpose uses unsupervised clustering we provide an approximate human description of the images corresponding to centroids detected using MMPose2D flow in the Table \ref{table:centroiddescription} below. This is curated by manually inspecting many batches of images labelled by the model. 

\begin{table}[H]
   \centering
   \caption{Approximate text description of cluster obtained}
   \label{table:centroiddescription}
   \begin{tabular}{ m{1cm} m{4cm}m{2cm} }
      \toprule
      \textbf{Centroid Label} & \textbf{Approx. Human Description} & \textbf{Image}\\ 
      \midrule
      0 & Upper body till waist $\mid$ nose visible $\mid$  front &
      \includegraphics[width=\linewidth]{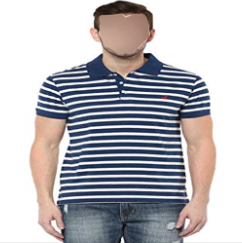}
      \\
      1 & Upper body till waist $\mid$  half $\mid$  back &
      \includegraphics[width=\linewidth]{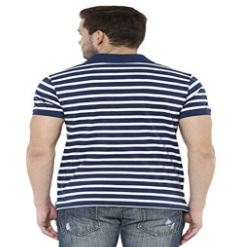}
      \\
      2 & Full body $\mid$ knee bent $\mid$  front & 
      \includegraphics[width=\linewidth]{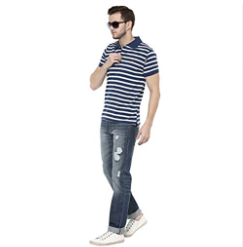}
      \\
      3 & Close up $\mid$  chin visible $\mid$  chest & 
      \includegraphics[width=\linewidth]{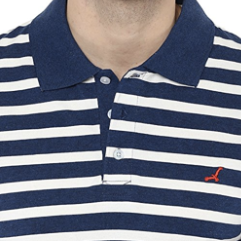}
      \\
      4 & Full body & 
      \includegraphics[width=\linewidth]{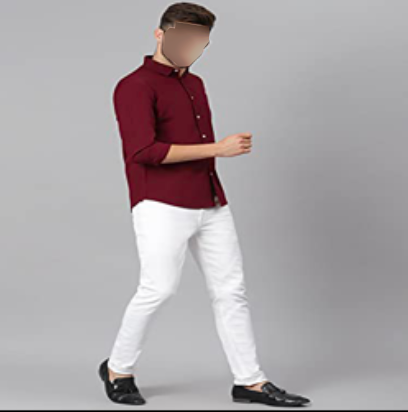}
      \\
      5 & Non-human $\mid$  tables $\mid$  fabric closeup& 
      \includegraphics[width=\linewidth]{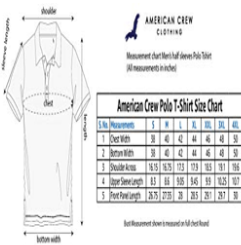}
      \\
      6 & Chin to torso& 
      \includegraphics[width=\linewidth]{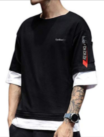}
      \\ 
      7 & Full body & 
      \includegraphics[width=\linewidth]{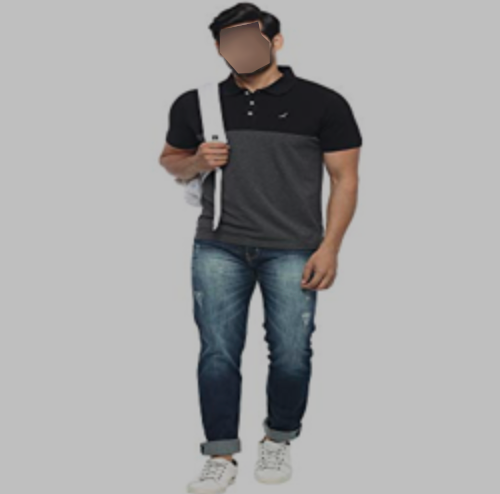}
      \\ 
      \bottomrule
   \end{tabular}

\end{table}

In the sections below we look at different subcategory of apparel and labels identified for prexisting images in the Product imageset and missing image labels proposed by the model.

\subsection{Polo T-Shirt}

\setcounter{figure}{6}
\begin{figure}[H]
   \centering
   \includegraphics[width=\linewidth]{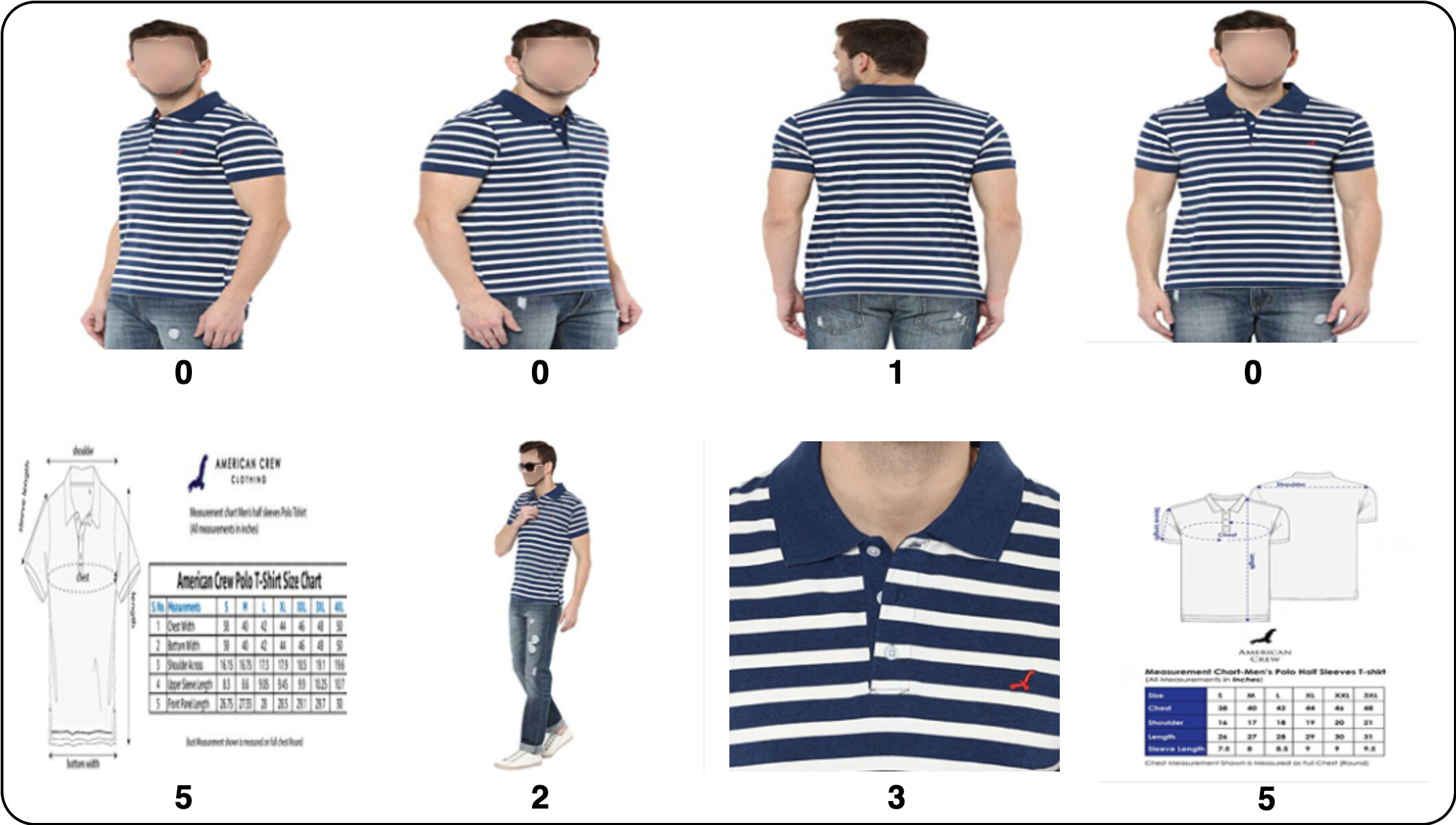}
 \caption{The figure shows the existing imageset for an Product and the labels predicted by the model. The numbers below the images correspond to the centroid index numbers from the Table \ref{table:centroiddescription}. The missing labels proposed by Unposed were:  {4, 6, 7} $\mid$ Refer Table \ref{table:centroiddescription}
}
   \label{fig:polotshirt}
\end{figure}

\subsection{T-Shirt}

\setcounter{figure}{7}
\begin{figure}[h]
   \centering
   \includegraphics[width=\linewidth]{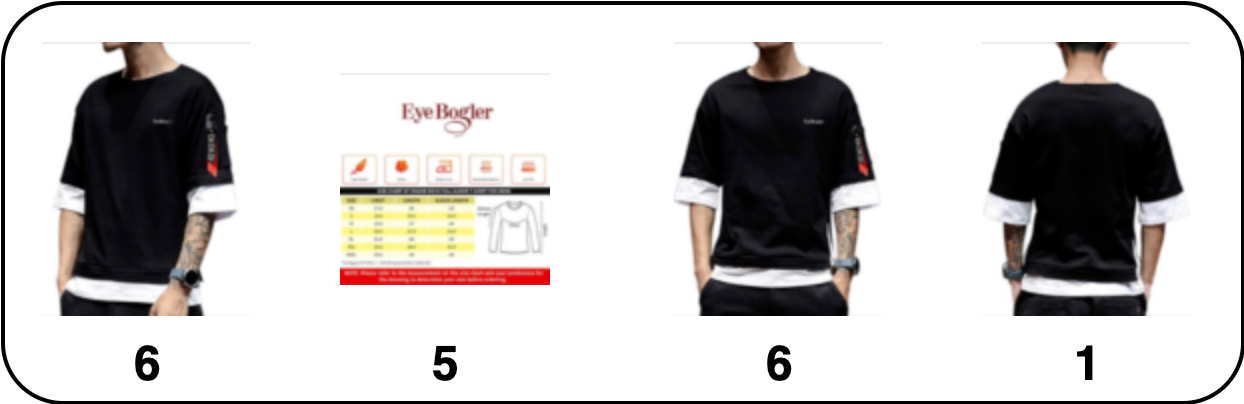}
\caption{The figure shows the existing imageset for an Product and the labels predicted by the model. The numbers below the images correspond to the centroid index numbers from the Table \ref{table:centroiddescription}. The missing labels proposed by Unposed were:    {0, 2, 3, 4, 7} $\mid$ Refer  Table \ref{table:centroiddescription}}
   \label{fig:tshirte1}
\end{figure}

\setcounter{figure}{8}
\begin{figure}[H]
   \centering
   \includegraphics[width=\linewidth]{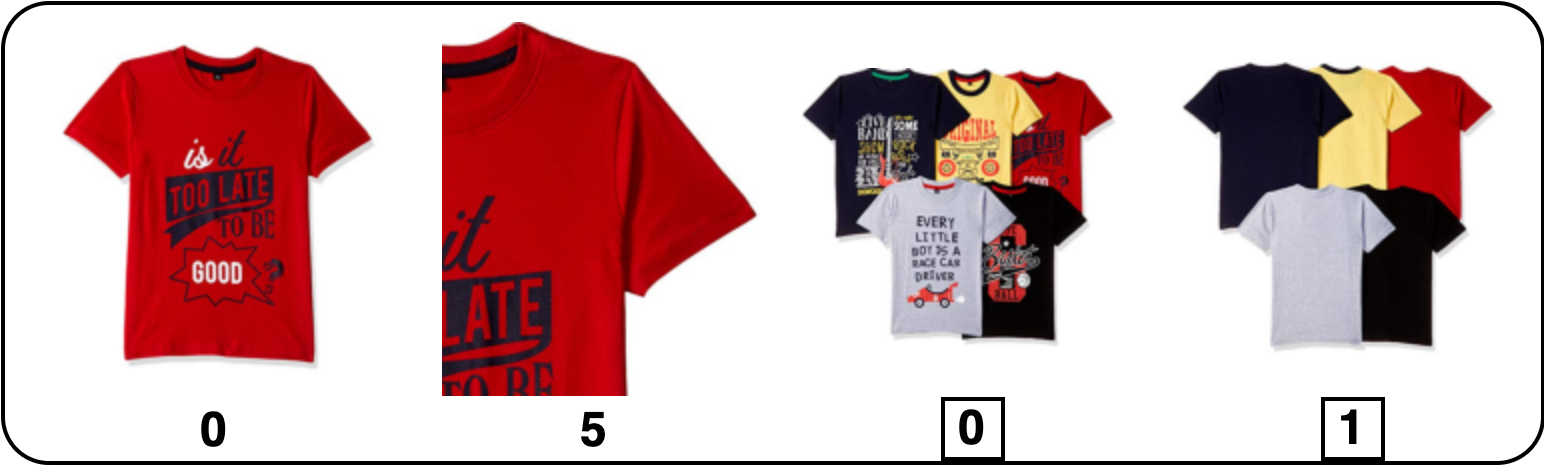}
   \caption{The figure shows the labels predicted by the MMPose model, the numbers under the images correspond to the centroid index numbers predicted for the existing imageset. The model predicts classes 0 and 1 corresponding to front and back poses as per Table \ref{table:centroiddescription}, though human keypoints are missing. This may acceptable for the use case as the desired poses are depicted. The resultant missing labels proposed by Unposed were:  {2, 3, 4, 7} $\mid$ Refer  Table \ref{table:centroiddescription} $\mid$ $\S$\ref{sec:notes}}
   \label{fig:tshirte2}
\end{figure}

\end{document}